%% file: arxiv.tex
\documentclass[10pt,twocolumn,letterpaper]{article}

\usepackage{iccv}
\usepackage{times}
\usepackage{epsfig}
\usepackage{graphicx}
\usepackage{amsmath}
\usepackage{amssymb}
\usepackage{booktabs}
\usepackage{bm}
\usepackage{algorithm}
\usepackage{algorithmicx}
\usepackage{xcolor}
\usepackage{color}
\usepackage{amstext}
\usepackage{algpseudocode}
\usepackage{multirow}
\usepackage{epstopdf}
\usepackage{booktabs,tabularx,colortbl,multirow,array,makecell}
\usepackage{pifont}
\usepackage{xr-hyper}
\usepackage{float}
\usepackage{caption}
\usepackage{subcaption}
\newcommand{\SO}[1]{\bm{\mathrm{SO}(#1)}}

\usepackage{abstract} 
\usepackage{lipsum} 

\usepackage[pagebackref=true,breaklinks=true,letterpaper=true,colorlinks,bookmarks=false]{hyperref}
\usepackage[capitalize]{cleveref}
\iccvfinalcopy 

\def\iccvPaperID{4200} 

\ificcvfinal\fi

\definecolor{vision}{rgb}{0.92,0.96,1.0}
\definecolor{state}{rgb}{0.89,0.98,0.89}
\newcommand{\state}{\rowcolor{state}}
\newcommand{\vision}{\rowcolor{vision}}

\begin{document}

\title{UniDexGrasp++: Improving Dexterous Grasping Policy Learning via Geometry-aware Curriculum and Iterative Generalist-Specialist Learning}


\author{
Weikang Wan \textsuperscript{1}\thanks{Equal contribution.}\quad 
Haoran Geng \textsuperscript{1,3}\footnotemark[1] \quad
Yun Liu \textsuperscript{2} \\
Zikang Shan \textsuperscript{1} \quad 
Yaodong Yang \textsuperscript{1,3} \quad 
Li Yi \textsuperscript{2} \quad
He Wang \textsuperscript{1}\thanks{Corresponding author.} \\
\textsuperscript{1}Peking University \quad
\textsuperscript{2}Tsinghua University  \\
\textsuperscript{3}Beijing Institute for General Artificial Intelligence     \quad
}

\twocolumn[{%
\renewcommand\twocolumn[1][]{#1}%
\maketitle
\begin{center}
    \centering
    \captionsetup{type=figure}
    \vspace{-10mm}
\includegraphics[width=\linewidth]{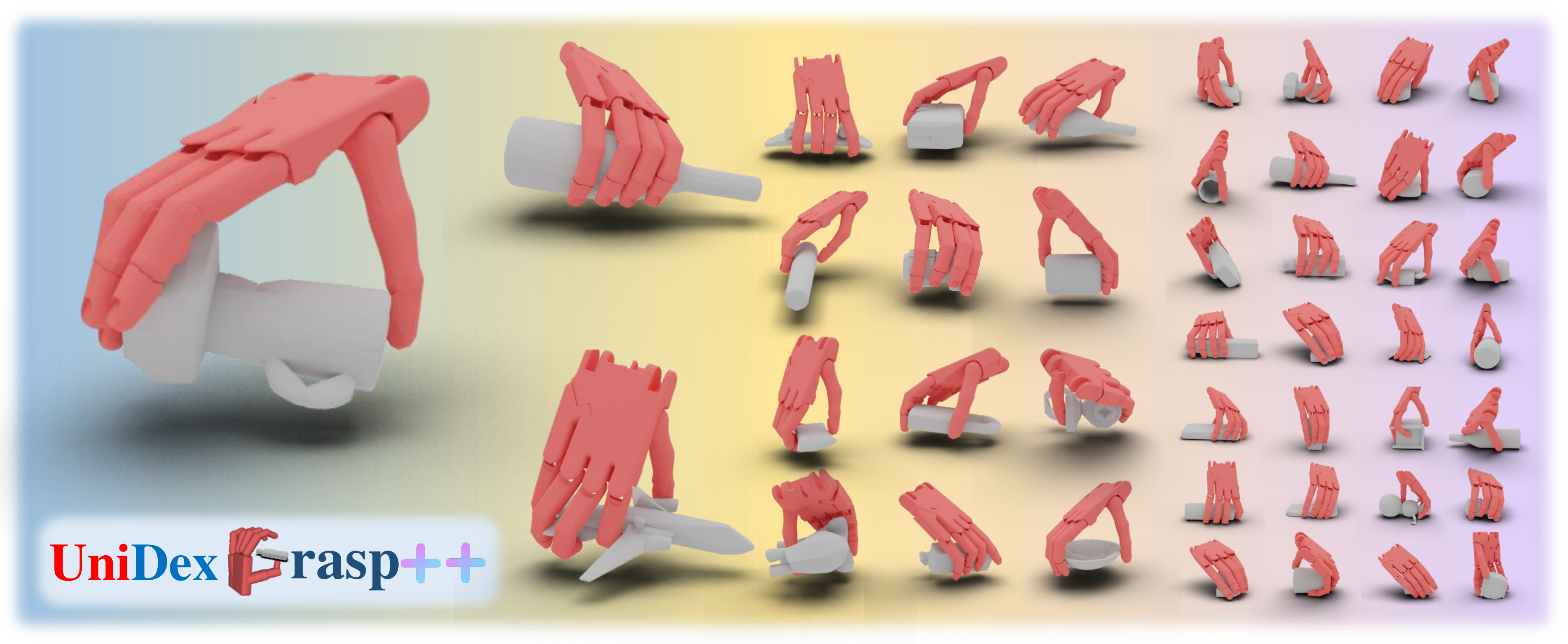}
\vspace{-8mm}
    \captionof{figure}{In this work, we present  a novel dexterous grasping policy learning pipeline, UniDexGrasp++.
    Same to UniDexGrasp\cite{xu2022universal},  UniDexGrasp++ is trained on 3000+ different object instances with random object poses under a table-top setting. It significantly outperforms the previous SOTA and achieves  85.4\% and 78.2\% success rates on the train and test set.}
    \label{fig:teaser}
\end{center}
}]

\ificcvfinal\thispagestyle{empty}\fi

\saythanks
\input{tex/0_abstract}
\input{tex/1_intro}

\input{tex/2_related_work}
\input{tex/3_problem_formulation}

\input{tex/4_method}
\input{tex/5_experiments}

\input{tex/6_conclusion}

{\small
\bibliographystyle{ieee_fullname}
\bibliography{egbib}
}

\newpage
\appendix
\input{tex/supp/00_method.tex}
\input{tex/supp/01_implementation.tex}
\input{tex/supp/02_additional_results}

\end{document}

%% file: tex/0_abstract.tex
\begin{abstract}
\vspace{-4mm}
We propose a novel, object-agnostic method for learning a universal policy for dexterous object grasping from realistic point cloud observations and proprioceptive information under a table-top setting, namely UniDexGrasp++. To address the challenge of learning the vision-based policy across thousands of object instances, we propose Geometry-aware Curriculum Learning (\textbf{\textit{GeoCurriculum}}) and Geometry-aware iterative Generalist-Specialist Learning (\textbf{\textit{GiGSL}}) which leverage the geometry feature of the task and significantly improve the generalizability. 
With our proposed techniques, our final policy shows universal dexterous grasping on thousands of object instances with \textbf{85.4\%} and \textbf{78.2\%} success rate on the train set and test set which outperforms the state-of-the-art baseline UniDexGrasp by 11.7\% and 11.3\%, respectively. 

\end{abstract}

%% file: tex/1_intro.tex
\section{Introduction}
\vspace{-2mm}
Robotic grasping is a fundamental and extensively studied problem in robotics, and it has recently gained broader attention from the computer vision community. Recent works~\cite{sundermeyer2021contact,breyer2021volumetric,fang2020graspnet,gou2021RGB,Wang_2021_ICCV,fang2022transcg,dai2022graspnerf} have made significant progress in developing grasping algorithms for parallel grippers, using either reinforcement learning or motion planning. However, traditional parallel grippers have limited flexibility, which hinders their ability to assist humans in daily life. 

Consequently, dexterous grasping is becoming more important, as it provides a more diverse range of grasping strategies and enables more advanced manipulation techniques.
The high dimensionality of the action space (e.g., 24 to 30 degrees of freedom) of a dexterous hand is a key advantage that provides it with high versatility and, at the same time, the primary cause of the difficulty in executing a successful grasp.
What's more, the complex hand articulation significantly degrades motion planning-based grasping methods, making RL the mainstream of dexterous grasping.

However, it is very challenging to directly train a vision-based universal dexterous grasping policy~\cite{mandikal2021dexvip,mandikal2021graff,mu2021maniskill,shen2022learning}. First, vision-based policy learning is known to be difficult, since the policy gradients from RL are usually too noisy to update the vision backbone. Second, such policy learning is in nature a multi-task RL problem that carries huge variations (e.g., different geometry and poses) and is known to be hard~\cite{mu2021maniskill,jia2022improving,shen2022learning}. Despite recent advancements in reinforcement learning (RL)~\cite{andrychowicz2020learning,akkaya2019solving,mandikal2021dexvip,chen2021system,christen2022dgrasp,rajeswaran2017learning,nagabandi2020deep,huang2021generalization,mandikal2021graff,she2022learning, wu2022learning}that have shown promising results in complex dexterous manipulation, the trained policy cannot easily generalize to a large number of objects and the unseen. At the same time, most works ~\cite{andrychowicz2020learning,akkaya2019solving,wu2022learning,christen2022dgrasp,she2022learning,rajeswaran2017learning,huang2021generalization} assume the robot knows all oracle information such as object position and rotation, making them unrealistic in the real world.

A recent work, UniDexGrasp~\cite{xu23UniDexGrasp}, shows promising results in vision-based dexterous grasping on their benchmark that covers more than 3000 object instances. Their policy only takes robot proprioceptive information and realistic point cloud observations as input.
To ease policy learning, UniDexGrasp proposes object curriculum learning that starts RL with one object and gradually incorporates similar objects from the same categories or similar categories into training to get a state-based teacher policy. After getting this teacher policy, they distill this policy to a vision-based policy using DAgger~\cite{ross2011reduction}. It finally achieves $73.7\%$ and $66.9\%$ success rates on the train and test splits. 
One limitation of UniDexGrasp is that its state-based teacher policy can only reach $79.4\%$ on the training set, which further constrains the performance of the vision-based student policy. Another limitation in the object curriculum is unawareness of object pose and reliance on category labels.

To overcome these limitations, we propose UniDexGrasp++, a novel pipeline that significantly improves the performance of UniDexGrasp.
First, to improve the performance of the state-based teacher policy, we first propose Geometry-aware Task Curriculum Learning \textbf{\textit{(GeoCurriculum)}} that measures the task similarity based on the geometry feature of the scene point cloud. 
To further improve the generalizability of the policy, we adopt the idea of \textit{generalist-specialist learning}~\cite{teh2017distral, mu2020refactoring, ghosh2017divide, jia2022improving} where a group of specialists is trained on the subset of the task space then distill to one generalist. We further propose Geometry-aware iterative Generalist-Specialist Learning \textbf{\textit{GiGSL}} where we use the geometry feature to decide which specialist handles which task and iteratively do distillation and fine-tuning. Our method yields the best-performing state-based policy, which achieves \textbf{87.9\%} and \textbf{83.7\%} success rate on the train set and test set. Then we distill the best-performing specialists to a vision-based generalist and do GiGSL again on vision-based policies until it reaches performance saturation. 
With our full pipeline, our final vision-based policy shows universal dexterous grasping on 3000+ object instances with \textbf{85.4\%} and \textbf{78.2\%} success rate on the train set and test set that remarkably outperforms the state-of-the-art baseline UniDexGrasp by 11.7\% and 11.3\%, respectively. The additional experiment on Meta-World~\cite{yu2020meta} further demonstrates the effectiveness of our method which outperforms the previous SOTA multi-task RL methods.


%% file: tex/2_related_work.tex
\vspace{-2mm}
\section{Related Work}
\vspace{-2mm}

\begin{figure*}[t]
    \centering
    \vspace{-0.5cm}
    \includegraphics[width=1.0\textwidth]{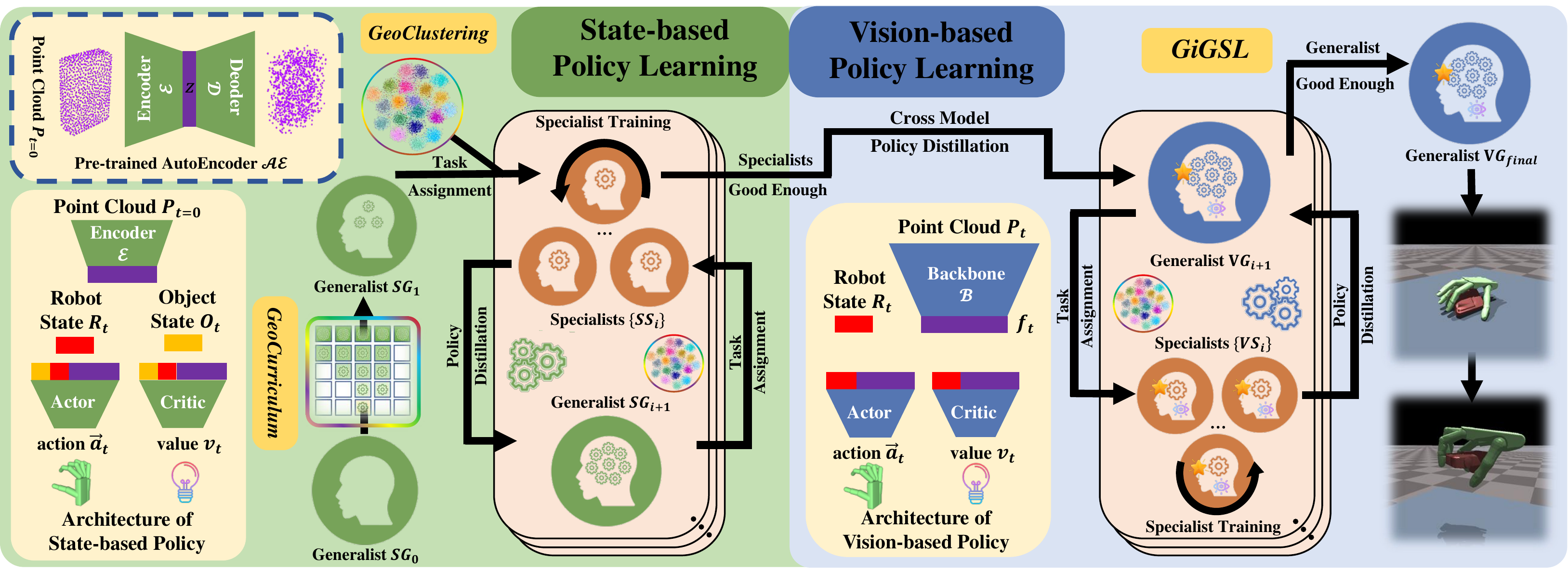}
    \vspace{-0.7cm}
    \caption{\textbf{Method Overview.} We propose to first adopt a state-based policy learning stage followed by a vision-based policy learning stage. The state-based policy takes input robot state $R_t$, object state $S_t$, and the geometric feature $z$ of the scene point cloud of the first frame. We leverage a geometry-aware task curriculum (\textit{\textbf{GeoCurriculum}}) to learn the first state-based generalist policy. After that, this generalist policy is further improved via iteratively performing specialist fine-tuning and distilling back to the generalist in our proposed geometry-aware iterative generalist-specialist learning (\textbf{\textit{GiGSL}}), where the task assignment to which specialist is decided by our geometry-aware
clustering (\textit{\textbf{GeoClustering}}). For vision-based policy learning, we first distill the final state-based specialists to an initial vision-based generalist and then do {\textit{GiGSL}} for the vision generalist, until we obtain the final vision-based generalist with the highest performance.
    }
    \vspace{-5mm}
    \label{fig:pipe}
\end{figure*}

\subsection{Dexterous Grasping}
\vspace{-2mm}
Dexterous hand has received extensive attention for its
potential for human-like manipulation in robotics~\cite{salisbury1982articulated,rus1999hand,okamura2000overview,dogar2010push,andrews2013goal,dafle2014extrinsic,kumar2016optimal,kumar2016learning,nagabandi2020deep,lu2020planning,qi2022hand,liu2022hoi4d}. It is of high potential yet very challenging due to its high dexterity. Dexterous grasping is a topic of much interest in this field.
Some works~\cite{baiyunfeiKarenLiu, pushgrasping, andrews2013goal} have leveraged analytical methods to model the kinematics and dynamics of both hands and objects, but they typically require simplifications, such as using simple finger and object geometries, to ensure the feasibility of the planning process.
Recent success has been shown in using reinforcement learning and imitation learning methods~\cite{chen2021system,christen2022dgrasp,rajeswaran2017learning,nagabandi2020deep,huang2021generalization,andrychowicz2020learning,she2022learning, wu2022learning}. While these works have shown encouraging results, they all suppose that the robot can get all the oracle states (e.g., object position, velocity) during training and testing. However, this state information can not be easily and accurately captured in the real world. To mitigate this issue, some works~\cite{mandikal2021graff,mandikal2021dexvip, qin2022dexpoint,xu23UniDexGrasp} consider a more realistic setting with the robot proprioception and RGB image or 3D scene point cloud as the input of the policy which can be captured more easily in the real world. 
Our work is more related to the recently proposed work UniDexGrasp~\cite{xu23UniDexGrasp} which learns a vision-based policy over 3000+ different objects. In this paper, we propose a novel pipeline that significantly improves the performance and generalization of UniDexGrasp, namely UniDexGrasp++.

\vspace{-2mm}
\subsection{Vision-based Policy Learning}
\vspace{-2mm}
Extensive research has been conducted to explore the learning of policies from visual inputs~\cite{zhang2015towards,kalashnikov2018scalable, srinivas2020curl, yarats2021improving, stooke2021decoupling, geng2022end,geng2022gapartnet, geng2023partmanip}. To ease the optimization and training process, some works have utilized a pre-trained vision model and frozen the backbone, as shown in works such as~\cite{seo2022reinforcement, radosavovic2022real, seo2022masked}. Others, such as~\cite{wu2022learning, wu2022vatmart}, have employed multi-stage training.
Our work is more related to ~\cite{chen2021system,chen2022visual,xu23UniDexGrasp}, who firstly train a state-based policy and then distill to a vision-based policy. Also, our work makes good use of the pre-training of vision-backbone in the loop of imitation (supervised) learning and reinforcement learning which enables us to train a generalizable policy under the vision-based setting.
\vspace{-1mm}
\subsection{Generalization in Imitation Learning and Policy Distillation}
\vspace{-1mm}
To generalize to large environment variations (e.g., object geometry, task semantics) in policy learning, previous works have used imitation learning including behavior cloning~\cite{torabi2018behavioral,kelly2019hg}, augmenting demonstrations to Reinforcement Learning~\cite{rajeswaran2017learning,wu2022learning,radosavovic2021state,shen2022learning,duan2016benchmarking} and Inverse Reinforcement Learning~\cite{ng2000algorithms,abbeel2004apprenticeship,ho2016generative,fu2017learning,liu2019state} to utilize the expert demonstrations or policies. Some works~\cite{teh2017distral, mu2020refactoring, ghosh2017divide, jia2022improving} have adopted the \textit{Generalist-Speciliast Learning} idea in which a group of specialists (teacher) is trained on a subset of the task space, and then distill to a
single generalist (student) in the whole task space using the above imitation learning and policy distillation methods.
While these works have made great progress on several benchmarks~\cite{mu2021maniskill,yu2020meta,james2020rlbench,urakami2019doorgym,cobbe2020leveraging}, they either do not realize the importance of how to divide the tasks or environment variations for specialists or focus on a different setting to our method. In this work, we leverage the geometry feature of the task in the specialists' division and curriculum learning which greatly improves the generalizability.


%% file: tex/3_problem_formulation.tex
\vspace{-2mm}
\section{Problem Formulation}

In this work, we focus on learning a universal policy for dexterous object grasping from realistic point cloud observations and proprioceptive information under a table-top setting, similar to~\cite{xu23UniDexGrasp,qin2022dexpoint}.

We learn such a universal policy from a diverse set of grasping tasks.
A grasping task is defined as $\tau= (o, R)$, where $o\in \mathbb{O}$ is an object instance from the object dataset $\mathbb{O}$, and $R\in \SO3$ is the initial 3D rotation of the object. 
To construct the environment, we randomly sample an object $o$, let it fall from a height, which randomly decides an initial pose, and then move the object center to the center of the table. We always initialize the dexterous hand at a fixed pose that is above the table center. The task is successful if the position difference between the object and the target is smaller than a threshold value. 
This is a multi-task policy learning setting and we require our learned policy to generalize well across diverse grasping tasks, \textit{e.g.}, across random initial  poses and thousands of objects including the unseen.

%% file: tex/4_method.tex
\vspace{-1mm}
\section{Method}
\vspace{-1mm}
This section presents a comprehensive description of our proposed method for solving complex tasks. In Sec.~\ref{sec:method_overview}, we provide an overview of our approach along with the training pipeline. Our proposed method leverages DAgger-based distillation and iterative Generalist-Specialist Learning (iGSL) strategy, which is explained in detail in Sec.~\ref{sec:method_iGSL}. Moreover, we introduce Geometry-aware Clustering to decide which specialist handles which task, achieving Geometry-aware iterative Generalist-Specialist Learning (GiGSL), which is presented in Sec.~\ref{sec:method_GiGSL}. In Sec.~\ref{sec:method_GeoCur}, we present a Geometry-aware Task Curriculum Learning approach for training the first state-based generalist policy.


\vspace{-1mm}

\subsection{Method Overview}
\label{sec:method_overview}
 Following ~\cite{xu23UniDexGrasp,chen2021system,chen2022visual},  we can divide our policy learning into two stages: 1) the state-based policy learning stage; 2) the vision-based policy learning stage. It is known that directly learning a vision-based policy is very challenging, we thus first learn a state-based policy that can access oracle information and let this policy help and ease the vision-based policy learning. The full pipeline is shown in Figure \ref{fig:pipe}.

\noindent\textbf{State-based policy learning stage.} 
The goal of this stage is to obtain a universal policy, or we call it a \textit{generalist}, that takes inputs from robot state $R_{t}$, object state $O_{t}$, and the scene point cloud $P_{t=0}$ at the first frame. Here the object point cloud is fused from multiple depth point clouds captured by multi-view depth cameras. And we include $P_{t=0}$ in the input to retain the scene geometry information and we use the encoder of a pre-trained point cloud autoencoder to extract its geometry feature. Note that at this point cloud encoder is frozen to make it as simple as possible, so it doesn't interfere with policy learning. We leave the visual processing of $P_{t}$ to the vision-based policy.

Although learning a state-based policy through reinforcement learning is more manageable than learning a vision-based policy, it is still very challenging to achieve a high success rate under such a diverse multi-task setting. We thus propose a geometry-aware curriculum learning \textit{\textbf{(GeoCurriculum)}} to ease the multi-task RL and improve the success rate.

After this \textbf{\textit{{GeoCurriculum}}}, we obtain the first state-based generalist $\text{SG}_1$ that can handle all tasks. We then propose a geometry-aware iterative Generalist-Specialist Learning strategy, dubbed as \textit{\textbf{GiGSL}}, to further improve the performance of the generalist. This process involves iterations between learning several state-based specialists $\{\text{SS}_i\}$ that specialize in a specific range of tasks and distilling the specialists to a generalist $\text{SG}_{i+1}$, where $i$ denotes the iteration index. The overall performance kept improving through this iterative learning until saturation.

\noindent\textbf{Vision-based policy learning.} For vision-based policy, we only allow it to access information available in the real world, including robot state $R_{t}$ and the scene point clouds $P_{t}$. In this stage, we need to jointly learn a vision backbone $\mathcal{B}$ that extracts $f_t$ from $P_{t}$
along with our policy (see the blue part of Fig.\ref{fig:pipe}). Here we adopt PointNet+Transformer~\cite{mu2021maniskill} as $\mathcal{B}$, which we find has a larger capacity and thus outperforms PointNet~\cite{qi2016pointnet}.
We randomly initialize the network weight of our first vision generalist $\text{VG}_1$. We start with performing a cross-modal distillation that distills the latest state-based specialists $\{\text{SS}_n\}$ to  $\text{VG}_1$.  
We can then start the \textit{GiGSL} cycles for vision-based policies that iterate between finetuning $\{\text{VS}_i\}$ and distilling to $\text{VG}_{i+1}$ until the performance of the vision-based generalist saturates. The final vision-based generalist $\text{VG}_\text{final}$is our learned universal grasping policy that yields the highest performance. Please refer to supplementary material for the pseudo-code of the whole pipeline.

\vspace{-1mm}
\subsection{\textit{\textbf{iGSL}}: iterative Generalist-Specialist Learning}
\vspace{-1mm}
 \label{sec:method_iGSL}
\noindent\textbf{Recap Generalist-Specialist Learning (\textit{\textbf{GSL})}}. The idea of Generalist-Specialist Learning comes from a series of works~\cite{teh2017distral, mu2020refactoring, ghosh2017divide, jia2022improving} that deal with multi-task policy learning. The most recent paper~\cite{jia2022improving} proposes \textit{{GSL}}, a method that splits the whole task space into multiple subspaces and lets one specialist take charge of one subspace. Since each subspace has fewer task variations and thus is easier to learn,  each specialist can be trained well and perform well on their task distributions. Finally, all the specialists will be distilled into one generalist.

Note that ~\cite{jia2022improving} only has one cycle of specialist learning and generalist learning. Straightforwardly, more cycles may be helpful. In \textit{{GSL}}, the distillation is implemented using GAIL~\cite{ho2016generative} or DAPG~\cite{rajeswaran2017learning} but we find their performance mediocre. In this work, we propose a better policy distillation method based on DAgger, iteratively enabling Generalist-Specialist Learning.

\noindent\textbf{Dagger-based policy distillation}. DAgger~\cite{ross2011reduction} is an on-policy imitation learning algorithm. Different from GAIL or DAPG, which only require expert demonstrations, DAgger~\cite{ross2011reduction} requires an expert policy, which is called a teacher, and the student that takes charge of interacting with the environment. When the student takes action, the teacher policy will use its action to serve as supervision to improve the student. Given that the student always uses its policy to interact with the environment, such imitation is on-policy and thus doesn't suffer from the covariate shift problem usually seen in the behavior cloning algorithm. Previous works, such as~\cite{xu23UniDexGrasp} for dexterous grasping and~\cite{chen2021system,chen2022visual} for in-hand manipulation, have used DAgger for policy distillation from a state-based teacher to a vision-based student and it is shown in UniDexGrasp~\cite{xu23UniDexGrasp} that DAgger outperforms GAIL and DAPG for policy distillation. 

However, one limitation of DAgger is that it only cares about the policy network and discards the value networks that popular actor-critic RL like PPO~\cite{schulman2017proximal} and SAC~\cite{haarnoja2018soft} usually have. In this case, when a teacher comes with both an actor and a critic distills to a student, the student will only have an actor without a critic and thus can't be further finetuned using actor-critic RL. This limits \textit{{GSL}} to simply one cycle and hinders it from further improving the generalist.

To mitigate this issue, we propose a new distillation method that jointly learns a critic function while learning the actor using DAgger.
Our DAgger-based distillation learns both a policy and a critic function during the supervised policy distillation process, where the policy loss is the mean squared error (MSE) between the actions from the teacher policy $\pi_{teacher}$ and the student policy $\pi_{\theta}$ (same in DAgger), and the critic loss is the MSE between the predicted value function $V_{\phi}$ and the estimated returns $\hat R_{t}$ using Generalized Advantage Estimation (GAE)~\cite{schulman2015high}.
\vspace{-2mm}
\begin{equation}
\vspace{-3mm}
\begin{aligned}  
    \mathcal{L} = \frac{1}{\left | D_{\pi_{\theta}} \right |} \sum_{\tau \in D_{\pi_{\theta}}}^{}(\pi_{teacher}(s_t)-\pi_{\theta}(s_t))^2  +\\
    \frac{1}{\left | D_{\pi_{\theta}} \right |T } \sum_{\tau \in D_{\pi_{\theta}}}^{}\sum_{t = 0}^{T}(V_{\phi}(s_t)-\hat R_{t})^2
\end{aligned}  
\vspace{-0.mm}
\end{equation}

This DAgger-based distillation method allows us to retain both the actor and critic while achieving very high performance. Compared to ILAD~\cite{wu2022learning} that only pre-trains the actor and directly finetunes the actor-critic RL (the critic network is trained from scratch), 
our method enables actor-critic RL to fine-tune 
on both trained actor and critic networks, enhancing the stability and effectiveness of RL training.

\noindent\textbf{Iteration between specialist fine-tuning and generalist distillation}. With our proposed DAgger-based distillation method, we can do the following: 1) start with our first generalist learned through \textit{{GeoCurriculum}}; 2) clone the generalist to several specialists, finetune each specialist on their own task distribution; 3) using DAgger-based distillation method to distill all specialists to one generalist; we can iterate between 2) and 3) until the performance saturates.

\vspace{-2mm}
\subsection{\textit{\textbf{GiGSL}}: Geometry-aware iterative Generalist-Specialist Learning}
\vspace{-2mm}
\label{sec:method_GiGSL}
\label{sec:geocluster}

One important question left for \textit{iGSL} is how to partition the task space. In ~\cite{jia2022improving}, they are dealing with a limited amount of tasks and it is possible for them to assign one specialist to one task or randomly. However, in our work, we are dealing with an infinite number of tasks considering the initial object pose can change continuously. We can only afford a finite number of specialists and need to find a way to assign a sampled task to a specialist. We argue that similar tasks need to be assigned to the same specialist since one specialist will improve effectively via reinforcement learning only if its task variation is small.
To this end, we propose \textit{\textbf{GeoClustering}}, a strategy for geometry-aware clustering in the task space.


\noindent\textbf{\textit{GeoClustering }strategy}. We split the task space $\mathbb{T} = \mathbb{O}\times \SO3$ into $N_\text{clu}$ clusters, with tasks in each cluster ${C}_j$ being handled by a designated specialist $S_j$ during specialist fine-tuning. We begin by sampling a large number of tasks $\{\tau^{(k)}\}_{k=1}^{N_\text{sample}}$ from $\mathbb{T}$ ($N_\text{sample}\approx 270,000$  in our implementation) and clustering their visual features using K-Means. The clustering of the large-scale task samples provides an approximation of the clustering of the whole continuous task space. 

We first train a point cloud 3D autoencoder  using the point cloud $\{P_{t=0}^{(k)}\}_{k=1}^{N\text{sample}}$ of the initialized objects in the sample tasks $\{\tau^{(k)}\}_{k=1}^{N\text{sample}}$. The autoencoder follows an encoder-decoder structure. The encoder $\mathcal{E}$ encodes $P_{t=0}^{(k)}$ and outputs the encoding latent feature $z^{(k)} = \mathcal{E}(P_{t=0}^{(k)})$. The decoder $\mathcal{D}$ takes $z^{(k)}$ as input and generates the point cloud $\hat{P}_{t=0}^{(k)}$.
The model is trained using the reconstruction loss $\mathcal{L}_\text{AE}$, which is the Chamfer Distance between ${P}_{t=0}^{(k)}$ and $\hat{P}_{t=0}^{(k)}$. See Supplementary Materials for more details.

During clustering for the state-based specialists, we use the pre-trained encoder $\mathcal{E}$ to encode the object point cloud $P_{t=0}^{(k)}$ for a task $\tau^{(k)}$ and obtain the latent code $z^{(k)}$. We use this geometry and pose encoded latent code $z^{(k)}$ as the feature for clustering. We then use K-Means to cluster the features of these sampled tasks $\{z^{(k)}\}_{k = 1}^{N_\text{sample}}$ and generate $N_\text{clu}$ clusters 
and corresponding cluster centers $\{c_j\}_{j=1}^{N_\text{clu}}$: 

And for vision-based specialists, thanks to the trained vision backbone, we  directly use it to generate feature $f^{(k)}$ to replace the corresponding encoding feature $z^{(k)}$ in the state-based setting. 
Finally, the clustering for specialists can be formulated as: 

During the specialists fine-tuning, we assign a given task $\tau^{(k)}$ to the specialist in an online fashion to handle the infinite task space. During fine-tuning, we assign $\tau^{(k)}$ to $SS_j$ or $VS_j$ if the Specialist have the nearest center $c_j$ to the feature $z^{(k)}$. or $f^{(k)}$. Then each Specialist only needs to train on the assigned task set and distill their learned specific knowledge to the Generalist.

\noindent\textbf{Summary and Discussion.} \textit{\textbf{GeoClustering}} strategy resolves the problem of task space partition, allows one specialist to focus on concentrated task distribution, and thus facilitates the performance gain for each specialist. Please refer to Algorithm~\ref{alg:1} for the pseudo-code of \textit{{GeoClustering}}.

As a way to partition task space, our geometry-aware clustering is much more reasonable and effective than category label-based partition, based on the following reasons: 1) not every object instance has a category label; 2) considering the large intra-category geometry variations, not necessarily objects that belong to the same category would be taken care by the same specialist; 3) object pose can also affect grasping, which is completely ignored in category label based partition but is well captured by our method.


\begin{algorithm}[t]\small
  \centering
  \caption{\textbf{\textit{GeoClustering}}}
  \label{alg:1}
  \begin{algorithmic}[1]
  \Require Task Space $\mathbb{T}$, Encoder $\mathcal{E}$ from the pre-trained AutoEncoder or backbone $\mathcal{B}$ from the Vision Policy. Number of target clusters $N_\text{clu}$
    \State Sample $N_\text{sample}$ tasks $\{\tau^{(k)}\}_{j=1}^{N_\text{sample}}$ from $\mathbb{T}$
    \State Get features: 
    
        state-based: $\{z^{(k)}\}_{k=1}^{N_\text{sample}} \leftarrow \{\mathcal{E}(P_{t=0}^{(k)})\}_{k=1}^{N_\text{sample}}$ 

        vision-based: $\{f^{(k)}\}_{k=1}^{N_\text{sample}} \leftarrow \{\mathcal{B}(P_{t=0}^{(k)})\}_{k=1}^{N_\text{sample}}$ 
    \State Get cluster centers using K-Means: 
    
        state-based:\ \ $\{c_j\}_{j=1}^{N_\text{clu}} \leftarrow \text{K-Means} (\{z^{(k)}\})$ 

        vision-based: $\{c_j\}_{j=1}^{N_\text{clu}} \leftarrow \text{K-Means} (\{f^{(k)}\})$  \\

\Return Cluster centers
$\{c_j\}_{j=1}^{N_\text{clu}}$

           
  \end{algorithmic}
\vspace{-1mm}
\end{algorithm}
\vspace{-3mm}
\begin{algorithm}[t]
\vspace{-0mm}
\small
  \centering
  \caption{\textbf{\textit{GeoCurriculum}}}
  \label{alg:2}
  \begin{algorithmic}[1]
  \Require Task Space $\mathbb{T}$, $N_\text{train}$ tasks for training $\{\tau^{(k)}\}_{k=1}^{N_\text{train}} \subset \mathbb{T}$, $N_\text{level}$ hierarchical levels of curriculum learning and $N_\text{sub}$ sub-clusters for each level, Encoder $\mathcal{E}$ from the pre-trained AutoEncoder
    \State Get features from the encoder: $\{z^{(k)}\}_{k=1}^{N_\text{train}}$
    
    \State \textbf{Level $0$}: Find the center of the feature space $z_c \leftarrow \textit{\textbf{GeoClustering}}(N_\text{clu} = 1)$ and the task $\tau_c$ with features nearest to $z_c$, train $\mathcal{C}_0 = \{\tau_c\}$ $\ (\text{where }\|\mathcal{C}_0\| = 1)$.
  
    \For { Level $l$ in  $1, \dots, N_\text{level}-1$,}
    
    \State \textbf{Level $l$}: Split each cluster of the Level $l-1$ into  $N_\text{sub}$ sub-clusters. Find the $N_\text{sub}$  tasks with features nearest to each sub-cluster feature center and add these tasks to $\mathcal{C}_l$, train $\mathcal{C}_l \ (\text{where }\|\mathcal{C}_l\| = N_\text{sub}^l)$.

    \EndFor
    \State \textbf{Level $N_\text{level}$}: train $\mathcal{C}_{N_\text{level}} = \{\tau^{(k)}\}_{k=1}^{N_\text{train}} \  (\text{where }\|\mathcal{C}_{N_\text{level}}\| = N_\text{train})$ \\
  \Return $\{\mathcal{C}_{l}\}_{l=0}^{N_\text{level}}$
  \end{algorithmic}
  
\vspace{-1mm}
\end{algorithm}

\vspace{-0mm}



\subsection{\textbf{\textit{GeoCurriculum}: Geometry-aware Task Curriculum Learning}}
\vspace{-2mm}
\label{sec:method_GeoCur}
\noindent\textbf{Problems of \textit{GiGSL} from Scratch} For state-based policy learning, we in theory can start \textit{GiGSL} from scratch.
One straightforward way is to directly learn a generalist from scratch on the whole task space and then improve it following G-S-G-S-... steps. However, learning this first generalist directly on the whole task space using reinforcement learning would be very challenging, usually yielding a generalist with an unsatisfactory success rate. 

An alternative would be to first learn $N_\text{clu}$ specialist, distill to a generalist, and then follow S-G-S-G-... steps. However, this is still very suboptimal. Given the huge variations in our task space, we need $N_\text{clu} >> 1$ so that the task variations in each specialist are small enough to allow them effectively learn. This large number of specialists would be very costly for training. Furthermore, because each specialist is trained separately from scratch, their policy can be substantially different from each other, which may lead to new problems. Considering two tasks that are similar but assigned to different specialists (they are just around the boundary of the task subspace). Then, since the two specialists are trained independently, there is no guarantee  that the specialists will do similar things to these two similar tasks, which means the policy is discontinuous around the subspace boundary. During policy distillation, a generalist may get significantly different action supervision from different specialists for those ``boundary tasks". 
As a result, this discontinuity in policy may lead to difficulty in convergence and hurt the policy generalization toward unseen tasks.

\noindent\textbf{Recap Object Curriculum in UniDexGrasp} Following UniDexGrasp~\cite{xu23UniDexGrasp}, we consider leveraging curriculum learning to make the first generalist learning easier.  ~\cite{xu23UniDexGrasp} introduced an object curriculum: they start with training a policy using RL to grasp one object instance (this object may be in the different initial poses); once this policy is well trained, they increase the number of objects by incorporating several similar objects from the same category and then finetuning the policy using RL on the new collection of objects; then, they increase the number of objects again by taking all objects from the category and finetune the policy; finally, they expand the object range to all different kinds of categories in the whole training objects and finish the final fine-tuning. ~\cite{xu23UniDexGrasp} shows that object curriculum is crucial to their performance, improving the success rate of their state-based policy from 31\% to 74\% on training set.

\noindent\textbf{\textit{{GeoCurriculum.}}} One fundamental limitation in the object curriculum used in ~\cite{xu23UniDexGrasp} is unawareness of object pose and reliance on category labels. Similar to our argument in the discussion of  Sec.\ref{sec:method_GiGSL}, we propose to leverage geometric features to measure the similarity between tasks, rather than object identity and category label. We thus introduce \textit{{GeoCurriculum}}, a geometry-aware task curriculum that leverages hierarchical task space partition.


In detail, we design a $N_\text{level}$ task curriculum that assigns  tasks with increasing level of variations to policy learning and facilitate a step by step learning. 
As shown in Algorithm~\ref{alg:2}, we first find a task $\tau^{(k_c)}$ with the feature nearest to the feature center of all sampled tasks and train the policy (Level 0). Then iteratively, for level $l$, we split each cluster  in the previous level $l-1$ into $N_\text{sub}$ sub-clusters (30 in our implementation) based on geometry feature $z^{(k)}$  and find $N_\text{sub}$ corresponding centers. We then add tasks that have features nearest to these sub-centers to the currently assigned tasks $\mathcal{C}_{i-1}$. 
Finally, we get the hierarchical task groups for the curriculum, that is:
\vspace{-2mm}
\begin{equation}
\vspace{-2mm}
\begin{aligned}  
    \{\mathcal{C}_{l}\}_{l=0}^{N_\text{level}} = \textbf{\textit{GeoCurriculum}}(\mathbb{T})
\end{aligned}  
\end{equation}

\begin{table}
  \centering
  \vspace{-0.3cm}
  \begin{tabular}{@{}l|ccc}
    \toprule
    Model & Train(\%) & \multicolumn{2}{c}{Test(\%)}  \\ \cline{3-4} 
    & & \begin{tabular}[c]{@{}c@{}} Uns. Obj. \\ Seen Cat. \end{tabular} & Uns. Cat. \\
    \hline
    \hline
\state    PPO\cite{schulman2017proximal} & 24.3 & 20.9 & 17.2 \\
\state    DAPG\cite{rajeswaran2017learning} & 20.8 &  15.3 &  11.1 \\
\state    ILAD\cite{wu2022learning} & 31.9 & 26.4 & 23.1 \\
\state    GSL\cite{jia2022improving} & 57.3 & 54.1 & 50.9 \\
\state    UniDexGrasp\cite{xu23UniDexGrasp} & 79.4 & 74.3 & 70.8 \\
\state    Ours (state-based) &\textbf{87.9} & \textbf{84.3} & \textbf{83.1} \\
    \hline
\vision    PPO\cite{schulman2017proximal}+DAgger\cite{ross2011reduction} & 20.6 & 17.2 & 15.0 \\
\vision    DAPG\cite{rajeswaran2017learning}+DAgger & 17.9 & 15.2 & 13.9 \\
\vision    ILAD\cite{wu2022learning}+DAgger & 27.6 & 23.2 & 20.0 \\
\vision    GSL\cite{jia2022improving}+DAgger & 54.1 & 50.2 & 44.8 \\
\vision    UniDexGrasp\cite{xu23UniDexGrasp} & 73.7 & 68.6 & 65.1 \\
\vision    Ours (state)+DAgger & 77.4 & 72.6 & 68.8 \\
\vision    Ours (vision-based) & \textbf{85.4} & \textbf{79.6} & \textbf{76.7} \\
    \bottomrule
  \end{tabular}
  \vspace{-0.3cm}
  \caption{\textbf{The Average Success Rate of the Evaluated Objects on Both Training and Test Set}. For better clarity, we use green for the state-based policy and blue for the vision-based policy.}
  \label{tab:MainExp}
  \vspace{-5mm}
\end{table}

During training, we iteratively train the policy under each assigned task set. From tackling only one task in $\mathcal{C}_0$ to all the training tasks in $\mathcal{C}_{N_\text{level}}$, the policy grows up step by step and have better performance than directly training it under all tasks.



%% file: tex/5_experiments.tex
\vspace{-3mm}
\section{Experiment}
\vspace{-1mm}

\subsection{Experiment Setting}
We evaluate the effectiveness of our method in the challenging dexterous grasping benchmark UniDexGrasp~\cite{xu23UniDexGrasp} which is a recently proposed benchmark suite designated
for learning generalizable dexterous grasping.

UniDexGarsp contains 3165 different object instances spanning 133 categories. Since the ground-truth grasp pose generation for pretraining and point cloud rendering processes are very expensive for UniDexGrasp environments, we only consider the non-goal conditioned setting in UniDexGarsp which does not specify the grasping hand pose.
Each environment is randomly initialized with one object and its initial pose, and the environment consists of a panoramic 3D point cloud $P_t$ captured from the fixed cameras for vision-based policy learning.

For the network architecture, we use MLP with 4 hidden layers (1024,1024,512,512) for the policy network and value network in the state-based setting, and an additional PointNet+Transformer~\cite{mu2021maniskill} to encode the 3D scene point cloud input in the vision-based setting. We freeze the vision backbone during the vision-based specialist training. We use $K=N_{clu}=20$ in our experiments. Other detailed hyperparameters are shown in supplementary materials.


\subsection{Main Results}
\begin{figure}[t]
    \centering
    \vspace{-0.5cm}
    \includegraphics[width=0.45\textwidth]{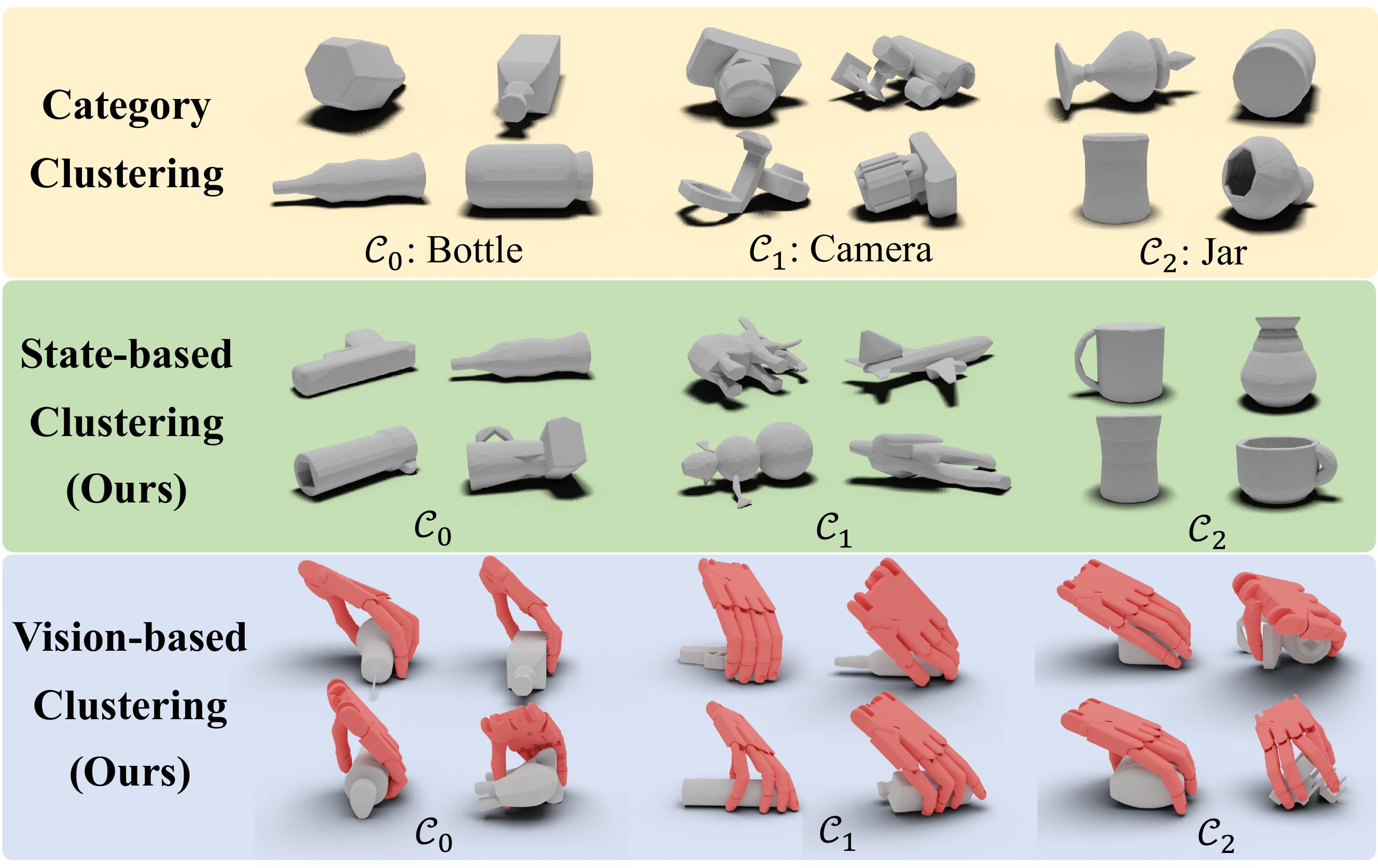}
    \vspace{-0.3cm}
    \caption{\textbf{Comparison between Category-label-based Clustering and our Geometry-aware Clustering.} Our state-based clustering is based on the features of the first-frame point clouds from the pre-trained encoder, while the vision-based policy utilizes its vision backbone to extract features for clustering. Due to the vision-based clustering being task-aware, we also show the grasping poses of the dexterous hands in the third row.}
    \label{fig:cluster}
     \vspace{-0.4cm}
\end{figure}

\begin{table*}[t]
\centering
\setlength\tabcolsep{2pt} 
\renewcommand{\arraystretch}{0.95}

\begin{tabular}{cc|cccccc|ccc}
\hline
& &\multicolumn{6}{c|}{Techniques}& \multicolumn{3}{c}{Success Rate (\%)}   \\ \hline
& &\begin{tabular}[c]{@{}c@{}}Geo-Aware \\ {Curri.} \end{tabular} &  \begin{tabular}[c]{@{}c@{}}Iterative \\ $\text{Fine-tuning}_\text{S}$ \end{tabular} & \begin{tabular}[c]{@{}c@{}}Geo-Aware \\ {Clustering} \end{tabular} &\begin{tabular}[c]{@{}c@{}}Iterative \\ $\text{Fine-tuning}_\text{V}$ \end{tabular} &  \begin{tabular}[c]{@{}c@{}}End2End \\ {Distillation} \end{tabular} & \begin{tabular}[c]{@{}c@{}}Transformer \\ {Backbone} \end{tabular}& \begin{tabular}[c]{@{}c@{}}Training \\ {\ } \end{tabular} & \begin{tabular}[c]{@{}c@{}}Test \\ Uns. Obj.  \end{tabular} &   \begin{tabular}[c]{@{}c@{}}Test \\  Uns. Cat. \end{tabular}  \\ \hline \hline
\state      &  \begin{tabular}[c]{|c}\ 1 \ \end{tabular}  &          &          &          &          &          &         &79.4 & 74.3 & 70.8 \\
\state  State    &  \begin{tabular}[c]{|c}\ 2 \ \end{tabular}  &\checkmark&          &          &          &          &         &82.7 & 76.8 & 74.2 \\   
\state  -based     & \begin{tabular}[c]{|c}\ 3 \ \end{tabular}   &\checkmark          &\checkmark&          &          &          &         &84.0 & 77.9 & 74.8 \\
\state    &  \begin{tabular}[c]{|c}\ 4 \ \end{tabular}  &\checkmark&\checkmark          &\checkmark&          &          &         &\textbf{87.9} & \textbf{84.3} & \textbf{83.1} \\   
\hline 


\vision     &  \begin{tabular}[c]{|c}\ 5 \ \end{tabular}  &\checkmark&\checkmark&&          &          &          &77.4 & 72.6 & 68.8 \\
\vision     &  \begin{tabular}[c]{|c}\ 6 \ \end{tabular}  &\checkmark&\checkmark&         & &\checkmark          &          &78.0 & 72.1 & 69.1 \\ 
\vision      &  \begin{tabular}{|c}\ 7 \ \end{tabular}  &\checkmark&\checkmark&          &\checkmark&\checkmark&          &78.9 & 74.7 & 70.2\\ 
\vision  Vision   &  \begin{tabular}{|c}\ 8 \ \end{tabular}  &\checkmark&\checkmark&\checkmark&\checkmark          &\checkmark&          &82.1 & 77.1 & 71.9 \\
\vision    -based &    \begin{tabular}[c]{|c}\ 9 \ \end{tabular} &\checkmark&\checkmark&\checkmark&\checkmark&          &\checkmark&82.7 & 76.2 & 73.4  \\
\vision     &    \begin{tabular}[c]{|c}10\end{tabular} &\checkmark&\checkmark&\checkmark& &\checkmark          &\checkmark&82.5 & 76.1 & 72.0 \\
\vision     &    \begin{tabular}[c]{|c}11\end{tabular} &\checkmark&\checkmark& &\checkmark &\checkmark          &\checkmark&78.6 & 73.7 & 72.3 \\
\vision     &  \begin{tabular}[c]{|c}12\end{tabular}  &\checkmark&\checkmark&\checkmark&\checkmark&\checkmark&\checkmark& \textbf{85.4} & \textbf{79.6} & \textbf{76.7} \\ \hline
\end{tabular}
\vspace{-3mm}
\caption{\textbf{Ablation Study.} For state-based policy (green) and vision-based policy learning (blue), we compare our techniques
with various ablations. }
\vspace{-5mm}
\label{table:Ablation}
\end{table*}

\begin{table}[t]
    \centering
    \begin{tabular}{c|ccc}
    \hline
          &PPO\cite{schulman2017proximal}& GSL\cite{jia2022improving}&Ours \\
         \hline
         MT-10 (\%)  & 58.4±10.1  
          & 77.5±2.9  
         & \textbf{80.3±0.5}  \\
         \hline
    \end{tabular}
    \vspace{-2mm}
    \caption{\textbf{Addtional Experiment in Meta-World.}}
    \label{tab:meta_world}
\vspace{-5mm}
\end{table}

We first train our method in the state-based policy learning setting and compare it with several baselines (green part in Tab.\ref{tab:MainExp}). We use PPO~\cite{schulman2017proximal} for the specialist RL in our pipeline. For these baselines: PPO~\cite{schulman2017proximal} is a popular RL method, DAPG~\cite{rajeswaran2017dapg}, and ILAD~\cite{wu2022learning} are imitation learning methods that further leverage expert demonstrations with RL; GSL~\cite{jia2022improving} adopts the idea of generalist-specialist learning  which use PPO for specialist learning and integrates demonstrations for generalist learning using DAPG, but with a random division for each specialist and only performs policy distillation once. UniDexGrasp~\cite{xu23UniDexGrasp} uses PPO and category-based object curriculum learning.
To compare our method to these baselines, we distill our final state-based specialists $\{\text{SS}_n\}$ to a state-based generalist $\text{SG}_{n+1}$ (although we won't use the latter later).
With our proposed techniques, our method achieves a success rate of 88\% and 84\% on the train and test set, which is \textbf{9\%} and \textbf{11\%} improvement over the UniDexGrasp in the state-based setting.

We then compare our method in the vision-based policy learning setting with the baseline methods(blue part in Tab.\ref{tab:MainExp}).  For PPO~\cite{schulman2017proximal}, DAPG~\cite{rajeswaran2017dapg}, ILAD~\cite{wu2022learning} and GSL~\cite{jia2022improving}, we distill the state-based policy to the vision-based policy using DAgger~\cite{ross2011reduction} since they don't consider the observation space change (state to vision) and directly training these methods under a vision input leads to completely fail. For our method, we compare our proposed whole pipeline ``Ours (vision-based)" with the variant of directly distilling our state-based policy to vision-based policy using DAgger, namely ``Ours (state)+DAgger". Our final results in the vision-based setting reach 85\% and 78\% on the train set and test set which outperforms the SOTA baseline UniDexGrasp for \textbf{12\%} and \textbf{11\%}, respectively.

\begin{figure}[t]
    \centering
    \vspace{-0.6cm}
    \includegraphics[width=0.46\textwidth]{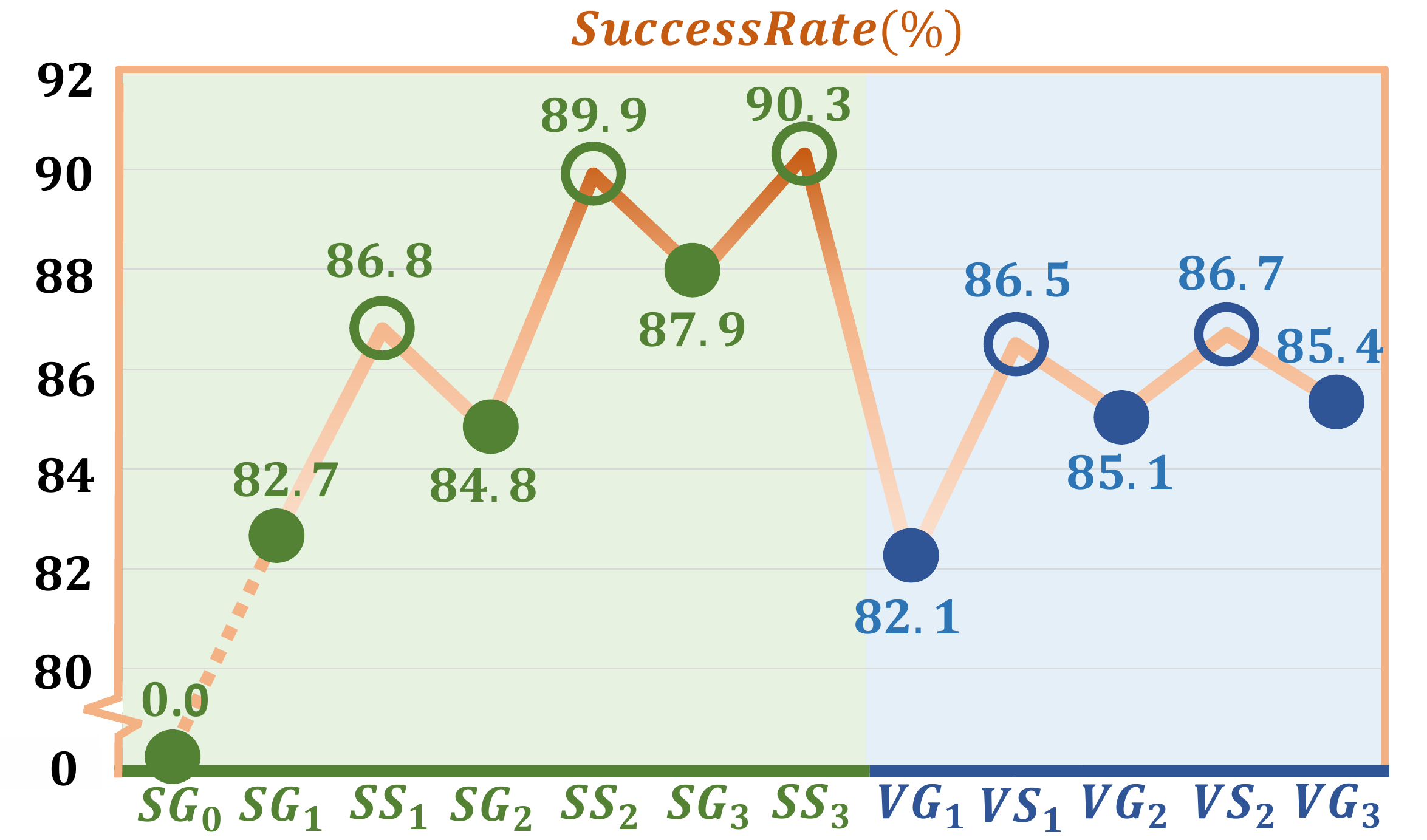}
    \vspace{-0.2cm}
    \caption{\textbf{Success Rate during our \textit{\textbf{GiGSL}} Training.} We plot the success rate of each training step, where green represents the state-based policy, blue represents the vision-based policy, hollow points represent the specialist policy, and solid points represent the generalist policy.}
    \label{fig:success}
    \vspace{-0.6cm}
\end{figure}

\vspace{-1mm}
\subsection{Analysis of the Training Process}
\vspace{-1mm}

\noindent \textbf{Geometry-aware Clustering Helps the Policy Learning.} We visualize some qualitative result in Fig.\ref{fig:cluster}. The first row shows a simple way of clustering, which is based on the object category. But as we analyzed above, this clustering method has no object geometry information and thus has limited help in grasping learning. The second row shows our  stated-based clustering strategy, which is based on the features from the point cloud encoder $\mathcal{E}$ and can cluster objects with similar shapes. And furthermore, in the third row, our vision-based clustering strategy utilizes the vision backbone which has more task-relative information, and thus the clustered objects have similar shapes as well as similar grasping poses.

\noindent \textbf{Quatitative Performance Improvement of our \textbf{\textit{GiGSL}}}. 
We visualize the success rate of each learning or fine-tuning step in Fig.\ref{fig:success}. No matter whether for state-based or vision-based policy, the improvement of Generalist-Specialist fine-tuning and distillation  shows the effectiveness of our Geometry-aware iterative Generalist-Specialist Learning
 \textit{\textbf{GiGSL}} strategy design and boosts the final performance of Universal Dexterous Grasping. 

\vspace{-2mm}
\subsection{Ablation Study}
\vspace{-1mm}
The ablation studies are shown in Tab.\ref{table:Ablation}. For \textbf{state-based policy learning stage} (green part), we analyze the ablation results as follows.

1) (Row 1,2) Effective of \textit{GeoCurriculum}. Using our proposed \textit{GeoCurriculum} (Row 2) performs better than using object-curriculum-learning in~\cite{xu23UniDexGrasp} (Row 1).

2) (Row 2,3) Effective of iterative $\text{fine-tuning}_\text{S}$. The policy can benefit from the iterative fine-tuning process and reach a higher success rate on both training and test set (Row 3) than a single cycle (Row 2).  Also see Figure \ref{fig:success}.

3) (Row 3,4) Effective of \textit{GeoClustering} in the state-based setting. With the pre-trained visual feature, the tasks assigned to one specialist are around the same feature clusters and thus are similar to each other.
This significantly reduces the difficulty of policy learning and, in return, improves performance (Row 4), compared to randomly assigning tasks to the specialists (Row 3).

For the ablation studies of \textbf{vision-based policy learning stage} (blue part), we use \textit{GeoClustering} in the state-based policy training by default, and the checkmark of \textit{GeoClustering} in this part indicates whether we use it in the vision-based policy learning. We use PointNet~\cite{qi2016pointnet} if there's no checkmark in ``Transformer Backbone".

4) (Row 5,6 \& 9,12) Effective of end-to-end distillation. We find directly distilling the final state-based specialists $\{\text{SS}_n\}$ to the vision-based generalist $\text{VG}_1$ (Row 6, 12) performs better than first distilling the state-based specialists $\{\text{SS}_n\}$ to the state-based generalist $\text{SG}_{n+1}$, then distilling this generalist to vision-based generalist $\text{VG}_1$ (Row 5, 9).

5) (Row 6,7 \& 10,12) Effective of iterative $\text{fine-tuning}_\text{V}$. The policy can benefit from the iterative fine-tuning process and reach a higher success rate on both training and test set than the single stage. Also see Figure \ref{fig:success}.

6) (Row 7,8 \& 11,12) Effective of \textit{GeoClustering} in the vision-based setting. By dividing the specialists using the learned visual feature from the vision backbone of the generalist, the final performance can be significantly improved than randomly dividing the specialists (\textbf{7\%} and \textbf{5\%} on training and test set, comparing Row 11 and 12).

7) (Row 8,12) Effective of the Transformer backbone.
The results show that the PointNet+Transformer backbone~\cite{mu2021maniskill} (Row 13) has a better expressive capacity which can improve the performance of DAgger-based distillation than using the PointNet~\cite{qi2016pointnet} backbone (Row 8).

\vspace{-1mm}
\subsection{Addtional Experiment in Meta-World}
\vspace{-2mm}

To further demonstrate the effectiveness of our proposed training strategy, we conduct more experiments in  meta-world benchmark~\cite{yu2020meta}, which focus more on state-based multi-task learning. We use our \textit{\textbf{iGSL}} method to handle the MT10 tasks and the results in Tab.\ref{tab:meta_world} show that our method outperforms the previous SOTA methods and demonstrates the advantages of our training strategy design, which enables iterative distillation and fine-tuning. More details and results can be found in the Supplementary Materials.

%% file: tex/6_conclusion.tex
\vspace{-2mm}
\section{Conclusions and Discussions}
\vspace{-2mm}
In this paper, we propose a novel pipeline, UniDexGrasp++, that significantly improves the performance and generalization of UniDexGrasp.
We believe such generalizability is also essential for Sim2Real transfer for real robot dexterous grasping.
The limitation is that we only tackle the dexterous grasping task in simulation and we will conduct the real-robot extension in our future work.

%% file: tex/supp/00_method.tex
\section{Method and Implementation Details}

\label{sec:method-detail}
We formalize our whole pipeline method in Algorithm~\ref{alg:1}.

\subsection{Details about Our Method}

\begin{algorithm}[h]\small
  \centering
  \caption{UniDexGrasp++}
  \label{alg:1}
  \begin{algorithmic}[1]
  \Require Task Space $\mathbb{T}$, $K$ State-based Specialists $\{SS^j\}$, a State-based Generalist $SG$, $K$ Vision-based Specialists $\{VS^j\}$, a Vision-based Generalist $VG$
    \State $\{\mathcal{C}_l\} \leftarrow$  \textit{\textbf{GeoCurriculum}}($\mathbb{T}$) for object curriculum.
    \State Geometry-aware task curriculum learning to train $SG_0$.
    \For {$i=1,2,\dots$}:
            \State Initialize specialist $SS_i^j = SG_i$ 
            \State $\{c_j\}\leftarrow $\textit{\textbf{GeoClustering}}($\mathbb{T}$)
            
            \State Online assign tasks that are nearest to $c_j$ to  specialist $SS_i^j$ and train $SS_i^j$ \Comment{RL}    
        \If  {$\{SS_i^j\}$ are optimal} \textbf{break}
        \Else 
            \State Distill $\{SS_i^j\}$ to $SG_{i+1}$ \Comment{DAgger-based Distillation} 
    \EndIf
    \EndFor
    
    \State Distill $\{SS_i^j\}$ to $VG_0$ \Comment{DAgger-based Distillation} 

    \For {$i=1,2,\dots$}:
            \State Initialize specialist $VS^j_i = VG_i$ 
            \State $\{c_j\}\leftarrow $ \textit{\textbf{GeoClustering}}($\mathbb{T}$) 
            
            \State Online assign tasks that are nearest to $c_j$ to  specialist $VS_i^j$ and train $VS_i^j$  \Comment{RL}
            \State Distill $\{VS_i^j\}_{i=1}^K$ to $VG_{i+1}$ \Comment{DAgger-based Distillation} 
        \If  { $VG_{i+1}$ is optimal} \textbf{break}
    \EndIf
    \EndFor

  \end{algorithmic}
\vspace{-1mm}
\end{algorithm}

\noindent\textbf{Details of \textit{GiGSL}:} 
During the state-based policy learning stage, we terminate training when the success rate of the current policy $SS_n$ is only marginally better than the previous policy $SS_{n-1}$ (by less than 0.5\%). At this point, we distill $SS_n$ to the vision-based policy. In the vision-based policy learning stage, we stop training when the success rate of the current policy $VG_n$ is only marginally better than the previous policy $VG_{n-1}$ (by less than 0.5\%). We then use $VG_n$ as our final policy.\\

\noindent\textbf{Details of \textit{AutoEncoder}:} 
We train the  point cloud 3D autoencoder  using the point cloud $\{P_{t=0}^{(k)}\}_{k=1}^{N\text{sample}}$ of the initialized objects in the sample tasks $\{\tau^{(k)}\}_{k=1}^{N\text{sample}}$. The autoencoder follows an encoder-decoder structure. The encoder $\mathcal{E}$ encodes $P_{t=0}^{(k)}$ and outputs the encoding latent feature $z^{(k)} = \mathcal{E}(P_{t=0}^{(k)})$. The decoder $\mathcal{D}$ takes $z^{(k)}$ as input and generates the point cloud $\hat{P}_{t=0}^{(k)}$.
\begin{equation}
\begin{aligned}  
    z^{(k)} = \mathcal{E}(P_{t=0}^{(k)})
\end{aligned}  
\end{equation} 
\begin{equation}
\begin{aligned}  
    \hat{P}_{t=0}^{(k)} = \mathcal{D}(z^{(k)})
\end{aligned}  
\end{equation} 
The model is trained using the reconstruction loss $\mathcal{L}_\text{AE}$, which is the Chamfer Distance between ${P}_{t=0}^{(k)}$ and $\hat{P}_{t=0}^{(k)}$. 
\begin{equation}
\begin{aligned}  
   \mathcal{L}_\text{AE} = \text{ChamferDistance}(P_{t=0}^{(k)}, \hat{P}_{t=0}^{(k)})
\end{aligned}  
\end{equation}

\noindent\textbf{Details of \textit{GeoCurriculum}:} 
In our implementation, we choose $N_\text{level} =4$ and use a 4-stage \textit{GeoCurriculum} to train, where the task number is 1-300-900-$N_\text{train}$. We also compare different $N_\text{level}$ and the result can be found in Sec.~\ref{sec:add-result}

\subsection{Details about Baselines}
\label{sec:details_baselines}

\noindent\textbf{PPO}
\quad Proximal Policy Optimization (PPO)~\cite{schulman2017proximal} is a popular model-free on-policy RL method. We adopt PPO as our RL baseline. \\

\noindent\textbf{DAPG}
\quad Demo Augmented Policy Gradient (DAPG)~\cite{rajeswaran2017learning} is a popular imitation learning (IL) method that leverages expert demonstrations to reduce sample complexity. Following the approach of ILAD~\cite{wu2022learning}, we generate demonstrations using motion planning. \\

\noindent\textbf{ILAD}
\quad ILAD~\cite{wu2022learning} is an imitation learning method that enhances the generalizability of DAPG. It introduces a novel imitation learning objective on top of DAPG, which jointly learns the geometric representation of the object using behavior cloning from the generated demonstrations during policy learning. We use the same generated demonstrations as in DAPG in this method.\\

\noindent\textbf{GSL}
\quad Generalist-Specialist Learning (GSL)~\cite{jia2022improving} is a three-stage learning method that first trains a generalist using RL on all environment variations, then fine-tunes a large population of specialists with weights cloned from the generalist, each trained using RL to master a selected small subset of variations. Finally, GSL uses these specialists to collect demonstrations and employs DAPG for the IL part to train a generalist. For a fair comparison, we adopt PPO~\cite{schulman2017proximal} for the RL part and DAPG~\cite{rajeswaran2017learning} for the IL part in our implementation.\\

\noindent\textbf{UniDexGrasp}
\quad UniDexGrasp~\cite{xu23UniDexGrasp} is a two-stage learning method. In the first state-based stage, they propose Object Curriculum Learning (OCL), which starts RL with one object and gradually incorporates similar objects from the same or similar categories into the training to obtain a state-based teacher policy. Once they obtain this teacher policy, they use DAgger~\cite{ross2011reduction} to distill it to a vision-based policy.\\

%% file: tex/supp/01_implementation.tex
\section{Experiment Details}
\label{sec:exp-details}

\quad As described in Sec.4, we use PPO~\cite{schulman2017proximal} in \textit{GeoCurriculum} learning stage to get the first generalist $SG_1$, and in specialist learning stage $\{\text{SS}_i\}$, $\{\text{VS}_i\}$ to train these specialist. In the generalist learning stages $SG_i (i>1)$ and $VG_i$, we employ our proposed DAgger-based policy distillation. Note that we freeze the vision-backbone in the $\{\text{VS}_i\}$ learning stage. \\

\subsection{Environment Setup}
\noindent\textbf{State Definition}\quad
\begin{table}
    \centering
    \begin{tabular}{@{}l|cc}
    \toprule
        Parameters & Description \\ 
        \hline \hline
        $\bm{q} \in \mathbb{R}^{18}$ & joint positions \\
        $\bm{\dot{q}} \in \mathbb{R}^{18}$  & joint velocities \\
        $\bm{\tau}_{\text{dof}} \in \mathbb{R}^{24}$  & dof force  \\
        $x_{\text{finger}} \in \mathbb{R}^{3\times5}$                          
          & fingertip position              \\
        $\alpha_{\text{finger}} \in \mathbb{R}^{4\times5}$                        & fingertip orientation           \\
        $\dot{x}_{\text{finger}} \in \mathbb{R}^{3\times5}$                    
          & fingertip linear velocities     \\
        $\omega_{\text{finger}} \in \mathbb{R}^{3\times5}$  
          & fingertip angular velocities    \\
        $F_{\text{finger}} \in \mathbb{R}^{3\times5}$  & fingertip force   \\
        $\tau_{\text{finger}} \in \mathbb{R}^{3\times5}$ & fingertip torque  \\
        $\bm{t} \in \mathbb{R}^3$  & hand root global transition          \\
        $R \in \mathbb{R}^{3\times3}$  & hand root global orientation         \\
        $\bm{a} \in \mathbb{R}^{24}$   & action \\
        \bottomrule
    \end{tabular}
    \caption{Robot state definition.}
    \label{tab:robot_state}
\end{table}
The full state of the state-based policy is denoted as ${\mathcal{S}}^{\mathcal{S}}_t=(R_t, O_t, P_{t=0})$, while the full state of the vision-based policy is represented as ${\mathcal{S}}^{\mathcal{V}}_t=(R_t, P_t)$. The robot state $R_r$ is detailed in Table \ref{tab:robot_state}, and the object oracle state $O_t$ includes the object pose (3 degrees of freedom for position and 9 degrees of freedom for rotation matrix), linear velocity, and angular velocity. To accelerate the training process, we sample only 1024 points from the object and the hand in the scene point cloud $P_t$.\\

\noindent\textbf{Action Space}\quad
The action space is the motor command of 24 actuators on the dexterous hand. The first 6 motors control the global position and orientation of the dexterous hand and the rest 18 motors control the fingers of the hand. We normalize the action range to $(-1,1)$ based on actuator specification.\\

\begin{figure}[h]
    \centering
    \includegraphics[width=0.8\linewidth]{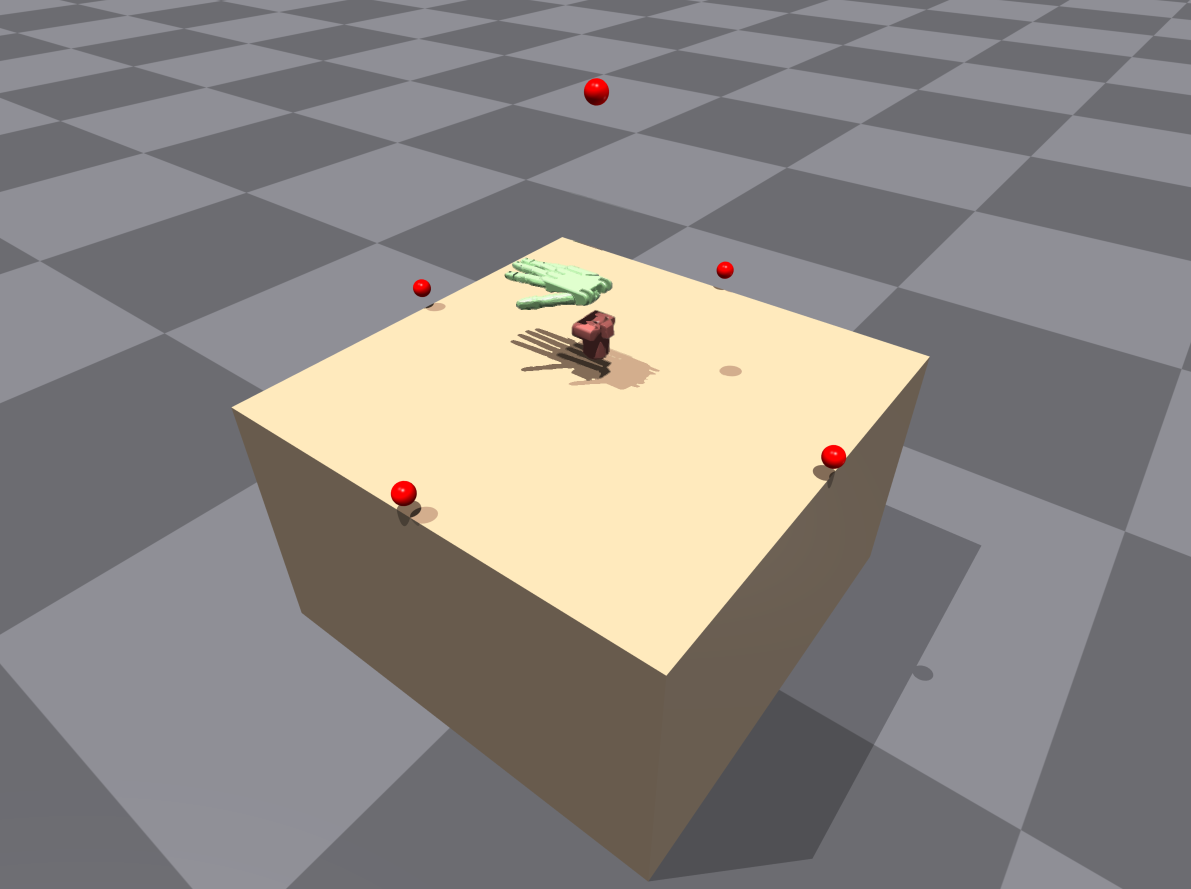}
    \caption{\textbf{Camera positions}}
    \label{fig:camera}
\end{figure}

\noindent\textbf{Camera Setup}\quad
Similar to UniDexGrasp~\cite{xu23UniDexGrasp}, we employ a setup consisting of five RGBD cameras positioned around and above the table, as shown in Fig. \ref{fig:camera}. The system's origin is located at the center of the table, and the cameras are positioned at ([0.5, 0, 0.05], [-0.5, 0, 0.05], [0, 0.5, 0.05], [0, -0.05, 0.05], [0, 0, 0.55]), with their focal points set to [0, 0, 0.05]. We fuse the partial point clouds generated by the five cameras to one scene point cloud $P_t$.\\

\noindent\textbf{Reward Function:}
\quad 
We use the non-goal-conditioned reward version in UniDexGrasp~\cite{xu23UniDexGrasp}, and we formalize it as follows (All the $\omega_{*}$ here are hyper-parameters same with UniDexGrasp.):

   
   

   The reaching reward $r_{\text{reach}}$ encourages the robot fingers to reach the object. Here, $\textbf{x}_{\text{finger}}$ and $\textbf{x}_{\text{obj}}$ denote the position of each finger and object:
   \begin{equation}
   r_{\text{reach}} = - \omega_{r} \sum{\lVert  \textbf{x}_{\text{finger}}-\textbf{x}_{\text{obj}} \rVert_2 }
   \end{equation}

   The lifting reward $r_{\text{lift}}$ encourages the robot hand to lift the object when the fingers are close enough to the object. $f$ is a flag to judge whether the robot reaches the lifting condition:
   $f=
   \textbf{Is} ( \sum{\lVert  \textbf{x}_{\text{finger}}-\textbf{x}_{\text{obj}} \rVert_2 }<\lambda_{f_1}) + \textbf{Is}({d}_{\text{obj}}>\lambda_{0})$. Here, ${d}_{\text{obj}} = \lVert  \textbf{x}_{\text{obj}}-\textbf{x}_{\text{target}} \rVert_2$, where $\textbf{x}_{\text{obj}}$ and $\textbf{x}_{\text{target}}$ are object position and target position. $a_z$ is the scaled force applied to the hand root along the z-axis ($\omega_{l}>0$).
   \begin{equation}
       r_{\text{lift}} = 
       \begin{cases}
       \omega_{l}*(1+a_z)  & \text{ if } f=2 \\
       0 & \text{ otherwise } 
       \end{cases}
   \end{equation}
    
   The moving reward $r_{\text{move}}$ encourages the object to reach the target and it will give a bonus term when the object is lifted very closely to the target:
   \begin{equation}
       r_{\text{move}} = 
       \begin{cases}
       -\omega_{m}{d}_{\text{obj}} + \frac{1}{1+\omega_{b}{d}_{obj}}  & \text{ if } {d}_{\text{obj}}<\lambda_{0} \\
       -\omega_{m}{d}_{\text{obj}} & \text{ otherwise } 
       \end{cases}
   \end{equation}
   
   Finally, we add each component and formulate our reward function as follows:
   \begin{equation}
   r = 
   r_{\text{reach}} +  r_{\text{lift}} +  r_{\text{move}}
   \end{equation} \\
   
\subsection{Training Details}

\noindent\textbf{Network Architecture:}
\quad The MLP used in the state-based policy ${\mathcal{\pi}}_{\mathcal{E}}$ and the vision-based policy  ${\mathcal{\pi}}_{\mathcal{S}}$ consists
of 4 hidden layers (1024, 1024, 512, 512). We use the exponential linear unit (ELU)~\cite{clevert2015fast} as the activation function.
The network structure of the 
PointNet in the autoencoder is (1024, 512, 64). 
We use the PointNet + Transformer backbone in~\cite{mu2021maniskill} as our vision backbone, where we use different PointNets~\cite{qi2016pointnet} to process points having different segmentation masks (robot, object, entire point cloud). There's also an additional MLP to output a 256-d hidden vector for the robot state alone. All the features from the MLP and PointNets are fed into a Transformer~\cite{vaswani2017attention}. The
output vectors are passed through global attention pooling to extract a representation of dimension 256, which is then provided into a final MLP with layer sizes [256, 128, feature\_dim] to output a visual feature, that is then concatenated with the robot state.

\noindent\textbf{Hyperparameters of Training:}
The hyperparameters in our experiments are listed in Tab.\ref{tab:SuppParam}.

\begin{table}
    \centering
    \begin{tabular}{@{}l|cc}
    \toprule
        Hyperparameter & Value  \\
        \hline \hline
        Num. envs (Isaac Gym, state-based) & 1024 \\
        Num. envs (Isaac Gym, vision-based) & 32 \\
        Env spacing (Isaac Gym) & 1.5\\
        Num. rollout steps per policy update (PPO) & 8 \\
        Num. rollout steps per policy update (DAgger) & 1 \\
        Num. batches per agent & 4 \\
        Num. learning epochs & 5 \\
        Buffer size (DAgger) & 2000 \\
        Episode length & 200 \\
        Saturation threshold of policy iteration & 0.005 \\
        Discount factor & 0.96 \\
        GAE parameter & 0.95 \\
        Entropy coeff. & 0.0 \\
        PPO clip range & 0.2 \\
        Learning rate & 0.0003 \\
        Value loss coeff. & 1.0 \\
        Max gradient norm & 1.0 \\
        Initial noise std. & 0.8 \\
        Desired KL & 0.16 \\
        Clip observations & 5.0 \\
        Clip actions & 1.0 \\
        $N_\text{sample}$ & 270,000 \\
        $N_\text{train}$ & 3200 \\
        $N_\text{clu}$ & 20 \\
        $\omega_{r}$ & 0.5 \\
        $\omega_{l}$ & 0.1 \\
        $\omega_{m}$ & 2 \\
        $\omega_{b}$ & 10 \\
        
        \bottomrule
         
    \end{tabular}
    \caption{\textbf{Hyperparameter for grasping policy.} }
    \label{tab:SuppParam}
\end{table}

\noindent\textbf{Training time:}
The experiment is done on four NVIDIA RTX 3090 Ti. The training process consists of 20,000 environment steps in the first stage of \textit{GeoCurriculum} and 15,000 environment steps (for every single policy) in other stages. It needs two days in total.

%% file: tex/supp/02_additional_results.tex
\section{Additional Results and Analysis}
This section contains extended results of the experiment depicted in Sec. 5. 

\label{sec:add-result}

\noindent\textbf{More ablation on \textit{GeoCurriculum}.}\quad
We do additional ablation experiments on the effectiveness of \textit{GeoCurriculum}, and the results are presented in Table \ref{tab:CurrAblation}. 
 Specifically, we compare our proposed \textit{GeoCurriculum} approach with not using any curriculum learning and with the object-curriculum-learning (OCL) method proposed in~\cite{xu23UniDexGrasp}. Our findings indicate that curriculum learning is essential for achieving success in the challenging dexterous grasping task with large variations in object instances and their initial poses. Moreover, we observed that our \textit{GeoCurriculum} approach, which considers the geometric similarity of different objects and poses, outperforms the OCL method, which only considers the category label of objects.
In addition, we do an ablation study on the number of curriculum learning stages. For the 3-stage \textit{GeoCurriculum}, the task number is 1-100-$N_\text{train}$; for the 4-stage \textit{GeoCurriculum}, the task number is 1-300-900-$N_\text{train}$; and for the 5-stage \textit{GeoCurriculum}, the task number is 1-20-100-1000-$N_\text{train}$. We compared the performance of $SG_1$ for all the experiments. Since the performance of the 5-stage \textit{GeoCurriculum} is similar to that of  the 4-stage \textit{GeoCurriculum}, we choose the 4-stage in our main experiment for simplicity.\\

\begin{table}[h]
  \centering
  \begin{tabular}{@{}l|ccc}
    \toprule
    Model & Train(\%) & \multicolumn{2}{c}{Test(\%)}  \\ \cline{3-4} 
    & & \begin{tabular}[c]{@{}c@{}} Uns. Obj. \\ Seen Cat. \end{tabular} & Uns. Cat. \\
    \hline
    \hline
No Curriculum &30.5 & 23.4 & 20.6 \\
OCL\cite{xu23UniDexGrasp} & 79.4 & 74.3 & 70.8 \\
\textit{GeoCurriculum} (3) & 81.3 & 75.6 & 73.3 \\
\textit{GeoCurriculum} (4) & 82.7 & \textbf{76.8} & \textbf{74.2} \\
\textit{GeoCurriculum} (5) & \textbf{82.9} & 76.4 & 74.0
\\
    \bottomrule
  \end{tabular}
  \caption{\textbf{Ablation study on \textit{GeoCurriculum}.} OCL refers to the Object Curriculum Learning proposed in~\cite{xu23UniDexGrasp}. The numbers in brackets represent the number of stages for curriculum learning.}
  \label{tab:CurrAblation}
\end{table}

\noindent\textbf{More ablation study on \textit{iGSL}}
For the policy distillation method used in iterative Generlist-Specilist Learning (\textit{iGSL}), we compare our DAgger-based policy distillation with several popular imitation learning methods, including Behavior Cloning (we also add a value function learning to make the process iterative), GAIL~\cite{ho2016generative} and DAPG~\cite{xu23UniDexGrasp}. We use \textit{GeoCurriculum} for all the methods and compare the performance of $SG_{n+1}$. Tab.\ref{tab:DaggerAblation} shows the results, which demonstrate that our DAgger-based policy distillation method significantly outperforms other methods. Notably, our method uses the teacher checkpoint, while other methods only use the demonstrations from the teacher.\\

\begin{table}[h]
  \centering
  \begin{tabular}{@{}l|ccc}
    \toprule
    Model & Train(\%) & \multicolumn{2}{c}{Test(\%)}  \\ \cline{3-4} 
    & & \begin{tabular}[c]{@{}c@{}} Uns. Obj. \\ Seen Cat. \end{tabular} & Uns. Cat. \\
    \hline
    \hline
BC + Value &12.4 & 8.6 & 8.4 \\
GAIL\cite{ho2016generative} &30.7 & 26.9 & 26.0 \\
DAPG\cite{xu23UniDexGrasp} & 61.4 & 52.6 & 47.9 \\
Ours &\textbf{87.9} & \textbf{84.3} & \textbf{83.1} \\
    \bottomrule
  \end{tabular}
  \caption{\textbf{Ablation study on the policy distillation method.} }
  \label{tab:DaggerAblation}
\end{table}

\noindent\textbf{More ablation study on \textit{GiGSL}}
We provide more ablation results of \textit{GiGSL}. First, we do ablation experiments on the cluster number $N_\text{clu}$ in \textit{GeoClustering}. We compare the performance of the final vision-based policy $VG_n$ for different $N_\text{clu}$. The results are in Tab.\ref{tab:ClusterAblation} which show that increasing  $N_\text{clu}$ beyond a certain point does not improve performance and may even decrease it. \\

\begin{table}[h]
  \centering
  \begin{tabular}{@{}l|ccc}
    \toprule
    Model & Train(\%) & \multicolumn{2}{c}{Test(\%)}  \\ \cline{3-4} 
    & & \begin{tabular}[c]{@{}c@{}} Uns. Obj. \\ Seen Cat. \end{tabular} & Uns. Cat. \\
    \hline
    \hline
0 (No specialist) &77.4 & 72.6 & 68.8 \\
10 &80.3 & 74.9 & 75.2 \\
20 &\textbf{85.4} & \textbf{79.6} & \textbf{76.7} \\
50 &77.2 & 71.2 & 69.9 \\
    \bottomrule
  \end{tabular}
  \caption{\textbf{Ablation study on the cluster number.} }
  \label{tab:ClusterAblation}
\end{table}

\noindent Then, we compare our \textit{GeoClustering} with random clustering and category label-based clustering (we evenly divide all the categories into $N_\text{clu}$ parts for a fair comparison). In category class-based clustering, we pre-train a classification task on all the objects and their initial poses. We then use the feature of the second-to-last layer for clustering and concatenate this feature to the robot state and object state in the state-based policy learning.
We compare the performance of the final vision-based policy $VG_n$ for different methods. The results are shown in Tab.\ref{tab:SpecialistAblation}.\\

\begin{table}[h]
  \centering
  \begin{tabular}{@{}l|ccc}
    \toprule
    Model & Train(\%) & \multicolumn{2}{c}{Test(\%)}  \\ \cline{3-4} 
    & & \begin{tabular}[c]{@{}c@{}} Uns. Obj. \\ Seen Cat. \end{tabular} & Uns. Cat. \\
    \hline
    \hline
Random &77.0 & 71.9 & 68.2 \\
Category Label. &79.7 & 73.9 & 74.1 \\
Ours &\textbf{85.4} & \textbf{79.6} & \textbf{76.7} \\
    \bottomrule
  \end{tabular}
  \caption{\textbf{Ablation study on the pre-trained autoencoder.} The features from the encoder are used in \textit{GeoClustering} in the state-based setting.}
  \label{tab:SpecialistAblation}
\end{table}

\noindent\textbf{More Results on Meta-World}
Here we show additional results on Meta-World~\cite{yu2020meta}, a popular multi-task policy learning benchmark. The MT-10 task consists of 10 diverse and challenging tasks, such as opening a door or picking up objects, that require a wide range of skills and abilities. The MT-50 task set is an extension of the MT-10 task set and includes 50 additional tasks that are even more complex and diverse. We
The results in Tab.\ref{tab:meta_world_more} demonstrate that our proposed technique, \textit{\textbf{iGSL}}, performs well on the multi-task MT-10 \& MT-50 and outperforms the baseline methods.\\

\begin{table}[h]
    \centering
    \begin{tabular}{c|ccc}
    \hline
          &PPO\cite{schulman2017proximal}& GSL\cite{jia2022improving}&Ours \\
         \hline
         MT-10 (\%)  & 58.4±10.1  
          & 77.5±2.9  
         & \textbf{80.3±0.5}  \\
          \hline
         MT-50 (\%)  & 31.1±4.5 
          & 43.5±2.2  
         & \textbf{45.9±1.7}  \\
         \hline
    \end{tabular}
    \caption{\textbf{Addtional Experiment in Meta-World.}}
    \label{tab:meta_world_more}
\end{table}

\begin{figure*}[t]
    \centering
    \includegraphics[width=0.8\linewidth]{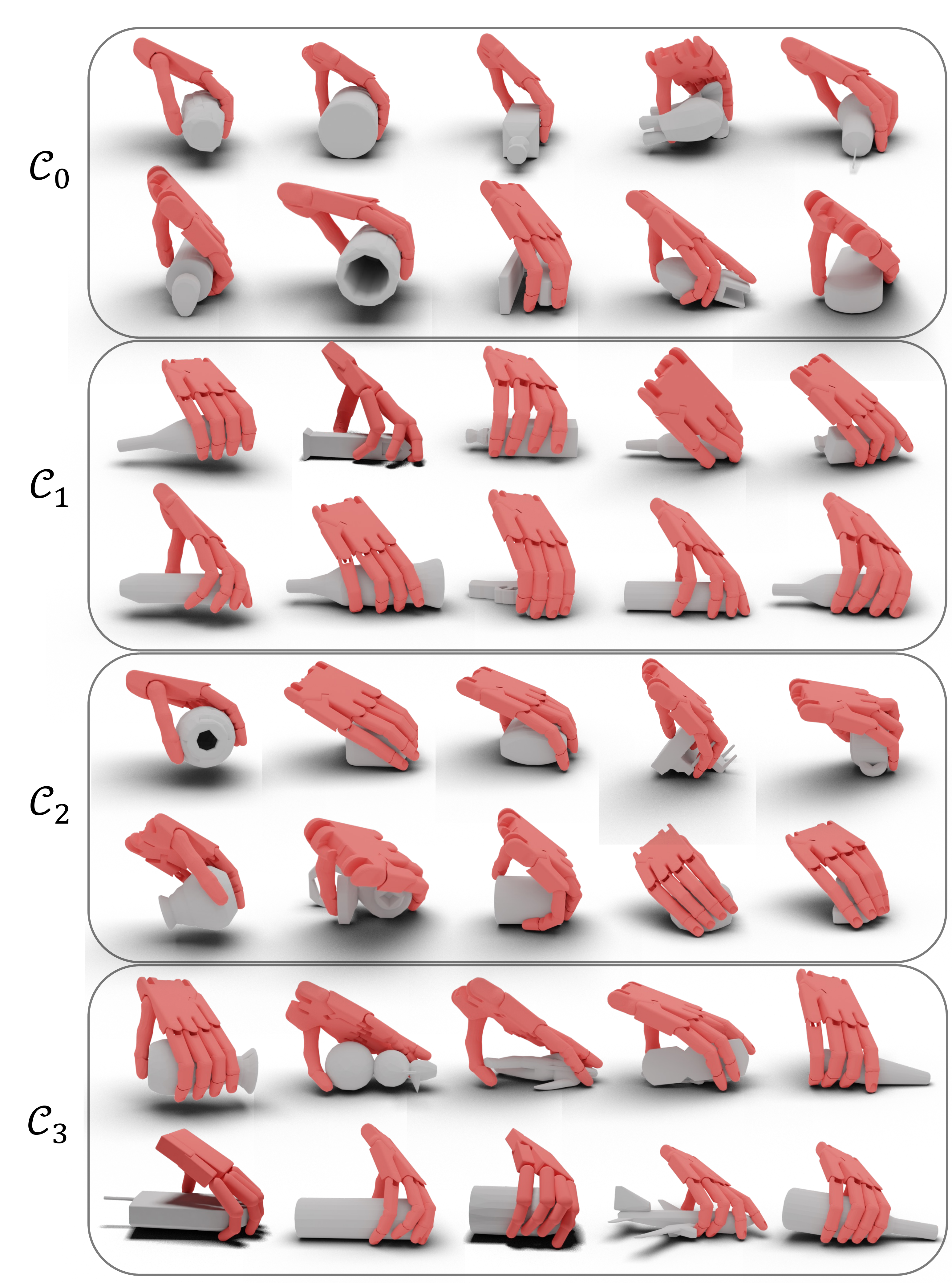}
    \caption{\textbf{Qualitative Grasping Results}. For each of the 4 clusters, we visualize 10 tasks and their corresponding grasping poses of the policy. The clusters are generated by our \textit{GeoClustering} in the vision-based policy learning stage. }
    \label{fig:gallery2}
\end{figure*}

\begin{figure*}[t]
    \centering
    \includegraphics[width=0.9\linewidth]{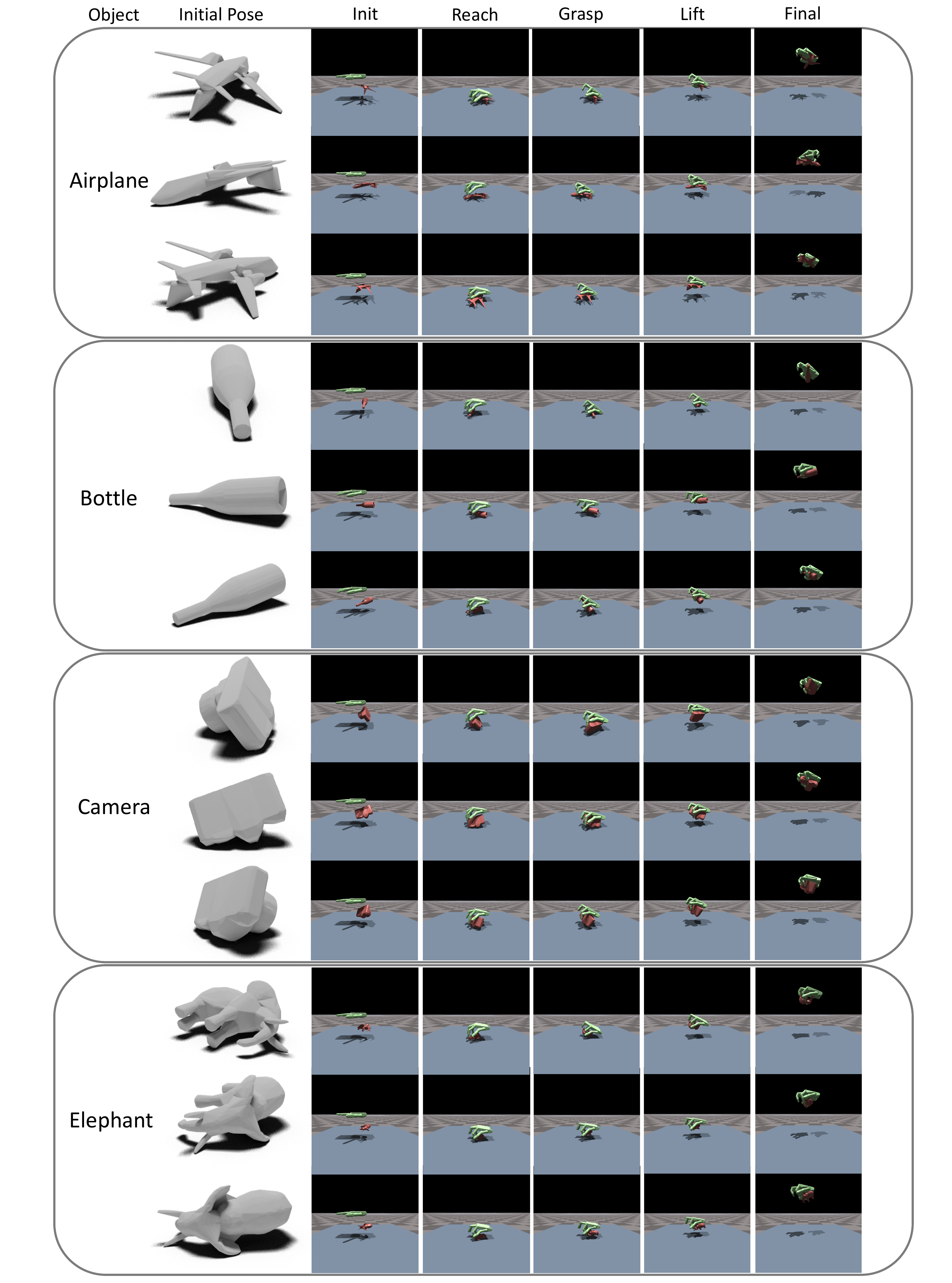}
    \caption{\textbf{Qualitative Grasping Trajecoties}. We provide several grasping trajectories for different objects with different initial poses.}
    \label{fig:gallery}
\end{figure*}

\noindent\textbf{Additional Qualitative Grasping Results}
We show more qualitative results in Fig.\ref{fig:gallery2} and  Fig.\ref{fig:gallery}.
In Fig.\ref{fig:gallery2}, we provide more results about our \textit{GeoClustering} in the vision-based policy learning stage. The vision-based policy $VG_1$ utilizes its vision backbone to extract visual features of the tasks for clustering. Due to the vision-based clustering being task-aware, we also show the grasping poses of these tasks. 
The results in Fig.\ref{fig:gallery2} demonstrate that our approach can cluster tasks based on the object geometry, pose features, and corresponding grasping strategy of the generalist policy.
In Fig.\ref{fig:gallery}, we provide several grasping trajectories for different objects with different initial poses.

